\pdfoutput=1

\documentclass[11pt]{article}

\usepackage[]{acl}

\usepackage{times}
\usepackage{latexsym}

\usepackage[T1]{fontenc}

\usepackage[utf8]{inputenc}

\usepackage{microtype}
\usepackage{algorithm}
\usepackage{algorithmic}
\usepackage{amssymb}
\usepackage{amsmath}
\usepackage{graphicx}
\usepackage{multirow}
\usepackage{stfloats}

\title{Improving the Adversarial Robustness of NLP Models \\ by Information Bottleneck}

\author{
Cenyuan Zhang$^{1,2}$\thanks{\ \ Equal contribution}, Xiang Zhou$^{1,2}$\footnotemark[1], Yixin Wan$^{3}$, \\  \textbf{Xiaoqing Zheng$^{1,2}$},
 \textbf{Kai-Wei Chang$^{3}$}, \textbf{Cho-Jui Hsieh$^{3}$}  \\
$^1$School of Computer Science, Fudan University, Shanghai, China \\
$^2$Shanghai Key Laboratory of Intelligent Information Processing \\
$^3$Department of Computer Science, University of California, Los Angeles, USA \\
\texttt{\{zhouxiang20,cenyuanzhang17,zhengxq\}@fudan.edu.cn} \\
\texttt{elaine1wan@g.ucla.edu},
\texttt{\{kwchang,chohsieh\}@cs.ucla.edu}
}

\begin{document}
\maketitle
\begin{abstract}
Existing studies have demonstrated that adversarial examples can be directly attributed to the presence of non-robust features, which are highly predictive, but can be easily manipulated by adversaries to fool NLP models.
In this study, we explore the feasibility of capturing task-specific robust features, while eliminating the non-robust ones by using the information bottleneck theory.
Through extensive experiments, we show that the models trained with our information bottleneck-based method are able to achieve a significant improvement in robust accuracy, exceeding performances of all the previously reported defense methods while suffering almost no performance drop in clean accuracy on SST-2, AGNEWS and IMDB datasets.
\end{abstract}

\section{Introduction}

Recently, a number of studies \cite{han-etal-2020-adversarial,aaai-20:Jin,DBLP:conf/iclr/ShafahiHSFG19} have revealed the fact that the performance of deep neural networks (DNNs) can be severely undermined by adversarial examples.
In the text domain, these adversarial examples are crafted by semantic-preserving perturbations to inputs with word synonym  substitution \cite{ebrahimi-etal-2018-hotflip, ren-etal-2019-generating,alzantot-etal-2018-generating} and  character-level transformations \cite{deepwordbug18, PSO20}. 
The vulnerability of DNN models results in inferior performances under adversarial attacks in many NLP tasks including text classification, natural language inference (NLI), question answering (QA), etc.
To resolve this problem, researchers have proposed various methods to defend against adversarial attacks \cite{goodfellow2014, szegedy-2013-Intriguing, jia-liang-2017-adversarial, kang-etal-2018-adventure,zhou-etal-2021-defense}.

In particular, \citet{Tsi-19} and \citet{advnotbugs-Ilyas} showed that the vulnerability of computer vision models can be attributed to ``non-robust features,'' which are features in the representation space that are sensitive to adversarial attacks and can be easily manipulated by attackers. The presence of these features will weaken the robustness of deep learning models. Therefore, a potential defense strategy is to filter out such non-robust features in the inputs. 

In this paper, we posit that the robustness of text classification models can be improved by filtering out the non-robust features. However, unlike the continuous input in the computer vision domain, input in the NLP domain is a sequence of words, which makes it difficult to model its distribution. We thus research into means to filter out the non-robust features in language model inputs.

Inspired by \cite{2019emnlp-li,wang2021infobert}, we propose to use the information bottleneck method \cite{arxiv-00:Tishby} in text classification tasks. Specifically, we plug in an information bottleneck layer (IB layer)\footnote{The source codes are available at \href{https://github.com/zhangcen456/IB}{https://github.com/\\zhangcen456/IB}.} between BERT \cite{bert} output layer and the text classifier to preserve only task-specific features. 

Since the IB layer trades off between minimizing preserved information and model prediction performance, features that are not robust under the targeted task are filtered out. Therefore, our approach is able to focus more on the robust features and achieve an improvement in its robustness.

We conduct extensive experiments on three text classification benchmarks: SST-2 \cite{sst-2}, AGNEWS \cite{agnews} and IMDB \cite{imdb}. Results have shown that our approach achieves a great improvement on model robustness compared with traditional defense methods, while only suffering little or even no performance drop on clean accuracy. We also provide a visualization to interpret how the information bottleneck layer works to keep robust features, in order to justify our proposed approach. 

In summary, we propose a new information bottleneck-based approach to improve the robustness of DNN language models. We demonstrate that our approach is effective in improving models' adversarial performance while maintaining their performance on clean data.
In addition, experimental results show that our approach can also be combined with existing adversarial training methods like FreeLB \cite{iclr2020-freelb} to further improve models' robustness under adversarial attacks.

\section{Related Work}

In response to the discovery of DNN's vulnerability to adversarial examples and the emergence of adversarial attacks, many defense methods have also been proposed to improve the robustness of DNN models. 

Among these methods, adversarial training is one of the most effective and widely used to defend against adversarial examples. In adversarial training, the model is trained to correctly classify both adversarial examples and normal examples. \citet{goodfellow2014} first propose a fast gradient sign method (FGSM) to generate adversarial examples for adversarial training in the image domain. In textual domain, many researchers tried to add perturbation to input word embedding to generate adversarial examples. \citet{zhang2018} applied several types of noises, such as Gaussian and Bernoulli, to perturb the input embeddings while \citet{iclr2020-freelb} proposed FreeLB, which minimizes the resultant adversarial loss inside different regions around input samples through adding adversarial perturbations to word embeddings. However, they all focus on the generalization of model, not the robustness.

\citet{wang2019} and \citet{Wang20} proposed to replace certain words in the training dataset with their synonyms for the purpose of data augmentation. However, this kind of methods are specific for the defense against synonym substitution attack, and may be weak when facing other kind of adversarial attack methods, such as character-level attack.

Apart from empirical methods, a set of certified robustness training methods is introduced recently, which have proven to be effective in improving a model's robustness against a specific type of attacks. \citet{Huang19} and \citet{jia-etal-2019-certified} used interval bound propagation (IBP) to propose certified robustness training methods that can limit the loss of the worst-case perturbations. These methods are provably robust to word substitution attacks. However, certified robustness training sacrifices the model's clean accuracy and is not generalized to all types of attacks. 

The information bottleneck method was proposed by \cite{arxiv-00:Tishby}, aiming to provide a quantitative notion of ``relevant information''. Although the method has been utilized in many NLP tasks such as rationale extraction~\cite{paranjape-etal-2020-information}, sentence summarization \cite{west-etal-2019-bottlesum} and parsing \cite{emnlp-20:Wang}, only few of them focus on combining information bottleneck with adversarial defense methods. \citet{iclr-21:Wang} proposes infoBERT, which applied information bottleneck to the embedding layer of pre-trained language models to suppress noisy information contained in word embeddings. The implicit assumption of this approach is that the embedding layer contains enough information for the model to make predictions. However, different word combinations can have different meanings, so the semantics of a sentence can not be fully expressed without taking contextual information into consideration.

Therefore, we propose a completely different implementation of the information bottleneck. The information bottleneck is utilized to extract task-related features from the output of the last layer of BERT, which is pre-trained and thus be capable of generating a contextualized representation for the input sequence, and calculated by using the variational inference method.

Besides, InfoBERT uses the gradient information of each word to find local anchored features and aims at increasing the mutual information between the global representation and them, while in our approach, robust features are extracted from the global representation without additional steps.

\section{Method}
\begin{figure}[ht]
\setlength{\belowcaptionskip}{-0.1cm}
    \centering
    \includegraphics[width=7.5cm]{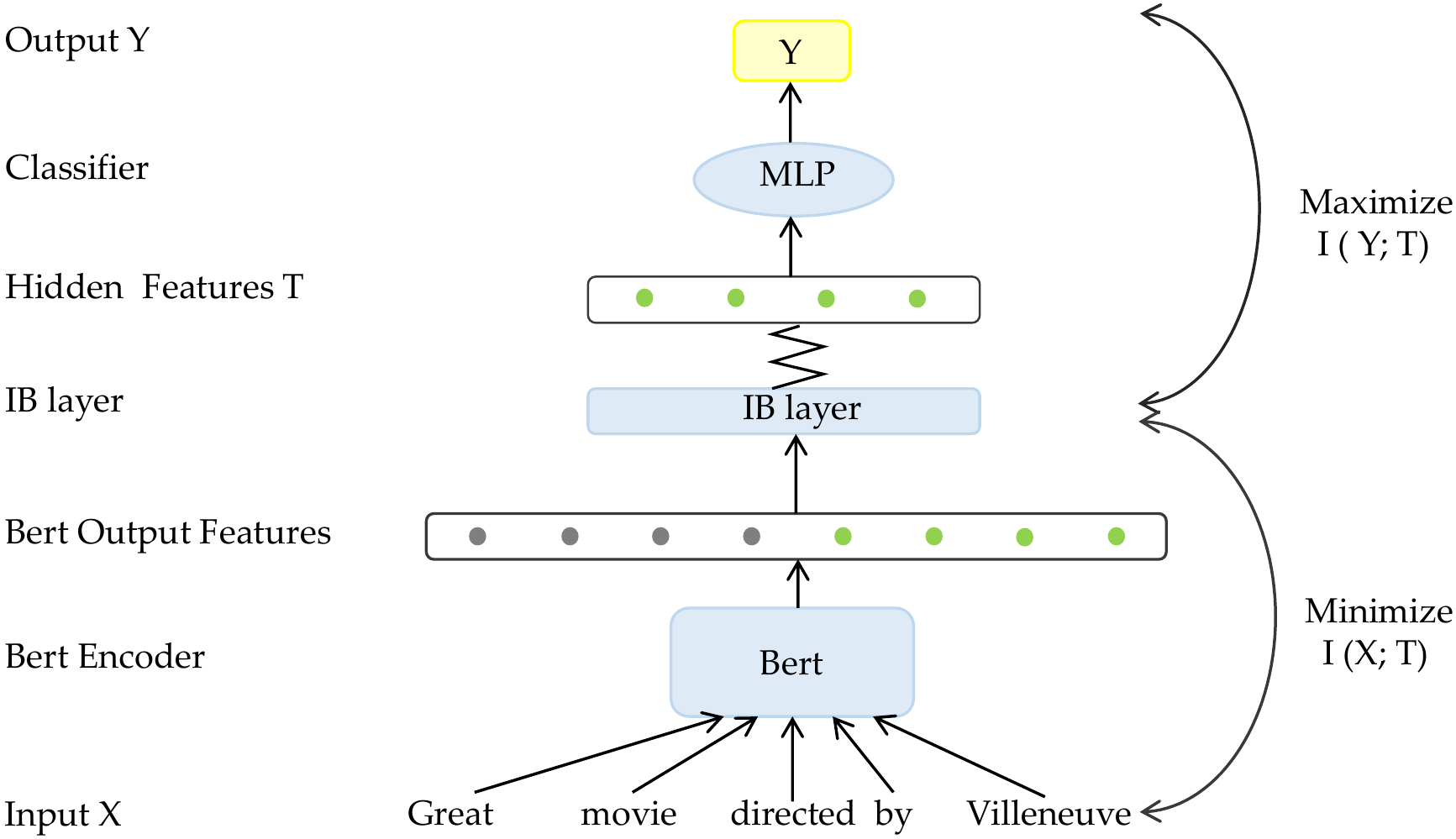}
    \caption{We use the IB layer to filter out non robust features, which are denoted by the gray circle dots in the figure, flow from Bert output. $I(\cdot;\cdot)$ denotes the mutual information and the jagged line denotes the compression process. By Maximizing the mutual information between the final prediction $\mathbf{Y}$ and the hidden features $\mathbf{T}$ while minimizing the mutual information between input $\mathbf{X}$ and $\mathbf{T}$ through IB layer, we are able to obtain a $\mathbf{T}$ that has all the non-robust features filtered out.}
    \label{fig:model}
\end{figure}
Derived from information theory, the information bottleneck method \cite{arxiv-00:Tishby} was proposed and has been used as a training objective as well as a theoretical framework \cite{ieee-15:Tishby} in  machine learning. 
The method of information bottleneck can be statistically formulated as follows: denote the input random variables as $\mathbf{X}$, which could be sentences or paragraphs, and the output as $\mathbf{Y}$. Denote the joint distribution of $\mathbf{X}$ and $\mathbf{Y}$ as $P(\mathbf{X},\mathbf{Y})$. The purpose of information bottleneck is to learn a distribution $p_\theta(t|x)$ from $\mathbf{X}$ to a compressed hidden feature $\mathbf{T}$. 
To simplify the notation, we will omit $\theta$ in
the subscript when we mention $p(t|x)$.
The information bottleneck objective (IB objective) used for optimization is as follows:
\begin{equation}
\mathcal{L}_{\text{IB}} = - I(\mathbf{Y};\mathbf{T}) + \beta \cdot I(\mathbf{X};\mathbf{T}),
\label{2}
\end{equation}

\noindent where $I(\cdot;\cdot)$ denotes the mutual information. The intuitive explanation for optimizing the information bottleneck objective Eq.\eqref{2} is that we want to compress all information given by the input $\mathbf{X}$, while still maintaining enough knowledge for the model to give the correct prediction outcome $\mathbf{Y}$. This can be achieved through finding the minimum value of $\mathcal{L}_{\text{IB}}$.
In the equation, parameter $\beta$ controls how much information we want to preserve among all the information extracted from the input $\mathbf{X}$. By increasing $\beta$, we can narrow the ``neck'', thus allowing less information from $\mathbf{X}$ to be transmitted to the hidden feature $\mathbf{T}$. 

Inspired by \cite{advnotbugs-Ilyas}'s theory about robustness of features, we  utilize the information bottleneck method to help the DNN models filter out ``non-robust features'' and only preserve ``robust features'' from model input. Since ''Robust features'' contribute to model's prediction, they contain semantic information of the input sequence. Taking this into account, our goal would be to filter out task-unrelated information while keeping the loss of task-related information to a minimum. 
This way, our method would be able to help improve model's robustness without diminishing its clean performance for the prediction task. By plugging in the IB layer right after BERT output, we can leverage the ability of pre-trained models on extracting contextualized features, preventing the possible loss of ``robust features'' after compression of information.

Specifically, given an input $\mathbf{X}$ and output $\mathbf{Y}$, we want to obtain specific hidden features $\mathbf{T}$ from $\mathbf{X}$ that only contain information which contributes to the final prediction $\mathbf{Y}$. By minimizing the IB objective in Eq.\eqref{2}, the IB layer filters out the task-unrelated information in BERT output, which is extracted from input $\mathbf{X}$, and obtain the required $\mathbf{T}$.

To minimize the IB objective, we maximize the mutual information $I(\mathbf{Y};\mathbf{T})$. Since the purpose of maximizing $I(\mathbf{Y};\mathbf{T})$ is to enforce $\mathbf{T}$ contain enough information for the model's prediction, we choose to minimize the loss function of the original task to approximate the maximization of $I(\mathbf{Y};\mathbf{T})$. Taking classification tasks as an example, our method would be to minimize the cross entropy function $\mathcal{L}_{\text{CE}}$.

The mutual information $I(\mathbf{X};\mathbf{T})$ can be calculated by the Kullback-Leibler distance between the distributions of $P(\mathbf{T}|\mathbf{X})$ and $P(\mathbf{T})$ as follows:

\begin{equation}
\begin{aligned}
I(\mathbf{X};\mathbf{T})=\mathbb{E}_X[D_{\text{KL}}[P(\mathbf{T}|\mathbf{X})||P(\mathbf{T})]]\\
= \int p(x,t)\log \frac{p(t|x)}{p(t)}dxdt.
\end{aligned}\label{5}
\end{equation}
To calculate the Kullback-Leibler divergence between $P(\mathbf{T}|\mathbf{X})$ and $P(\mathbf{T})$, we need knowledge of their probability distributions. The $P(\mathbf{T}|\mathbf{X})$ term can be sampled empirically. However, the $P(\mathbf{T})$ term is difficult to be estimated. To resolve this challenge, we expand the Eq.\ref{5} to get the following equation:
\begin{equation}
\begin{aligned}
I(\mathbf{X};\mathbf{T})= \int p(x,t) \log p(t|x) dxdt\\
- \int p(t)\log p(t) dt,
\end{aligned}
\end{equation}
where the marginal distribution of $\mathbf{T}$, $p(t) = \int p(t|x)p(x)dx$. Since the original \citet{arxiv-00:Tishby} relied on the iterative Blahut Arimoto algorithm to opitimize the IB objective, which is infeasible to apply to deep neural networks, many researchers try to use variational inference to approximate this problem \cite{information-dropout,DVIB017,IBG05}. Inspired by previous studies, we replace $p(t)$ with a variational approximation $q(t)=\mathcal{N}(\mu_X,\sigma_X^2)$,  which is a Gaussian distribution with the mean $\mu_X$ and standard deviation $\sigma_X^2$. Since the Kullback-Leibler divergence is defined to be non-negative, which means $\int p(t)\log p(t)dt \geq \int 
p(t)\log q(t)dt$, we derive the following upper bound:
\begin{equation}
\begin{aligned}
I(\mathbf{X};\mathbf{T}) \leq \int p(x)p(t|x)\log \frac{p(t|x)}{q(t)} dxdt\\
= \mathbb{E}_X[D_{\text{KL}}[P(\mathbf{T}|\mathbf{X})||Q(\mathbf{T})]].
\end{aligned}
\label{eq:upperbound}
\end{equation}

We want to reduce the mutual information between $\mathbf{X}$ and $\mathbf{T}$ so that more task-unrelated information can be filtered out, which can help us retain more robust features for the final prediction. 
To achieve this goal in practice, we minimize the upper bound of $I(\mathbf{X};\mathbf{T})$  derived in Eq.\eqref{eq:upperbound}. We achieve this through adjusting the parameters in $Q(\mathbf{T})$ in order to minimize the Kullback-Leibler divergence between $P(\mathbf{T}|\mathbf{X})$ and $Q(\mathbf{T})$, which will lower the upper bound of $I(\mathbf{X};\mathbf{T})$. Combined with the optimization goal of the term $I(\mathbf{Y};\mathbf{T})$ we explained in the former chapter, the final loss function is:
\begin{equation}
\mathcal{L} = \mathcal{L}_{\text{CE}} + \beta\cdot D_{\text{KL}}[P(\mathbf{T}|\mathbf{X})||Q(\mathbf{T})]\label{7}.
\end{equation}

By using the loss function Eq.\eqref{7} to optimize our model, our approach would be able to filter out the non-robust features for the classification task among all the inputs.

\section{Experiments}
In order to validate our assumption, we conduct several experiments to evaluate the effectiveness of our approach. We first compare our model and the baseline models both on their clean accuracy and accuracy under attack.
Furthermore, in exploration of the ability of our model to work in conjunction with adversarial training methods- such as FreeLB- to achieve complementary effects, we also evaluate the performance of the combined model. In addition, we try to interpret and further analyze our approach through several additional experiments.
\subsection{Dataset}
We evaluate our approach on three widely-used classification benchmark datasets: IMDB dataset \cite{imdb}, SST-2 \cite{sst-2} dataset, and AGNEWS dataset \cite{agnews}. Both IMDB and SST-2 are sentimental classification datasets with two classes, while AGNEWS is a topic classification dataset with four classes. 

\subsection{Baseline Models}
Because our model can be viewed as an enhanced variant of the BERT models, we choose to use BERT base \cite{bert} as one of the baseline models. 
We also establish a comparison with InfoBERT \cite{wang2021infobert}, a method that is very similar to our approach, to verify the effectiveness of the proposed way of injecting an IB layer.

Apart from these two baseline models, we also compare our approach with four adversarial training methods: PGD \cite{iclr2018-aleksander}, which is a classic and representative method, as well as FreeLB \cite{iclr2020-freelb}, SMART \cite{jiang-etal-2020-smart} and TAVAT \cite{tavat2021}, which are three state-of-the-art defense methods.
\subsection{Robustness Evaluation}

We evaluate models' accuracy under four different attack algorithms, including both word-level attacks and character-level attacks.

\noindent\textbf{Textfooler} \cite{jin2019textfooler} Textfooler ranks the importance of words by the drop of true class probability  after deleting words from the original text. By leveraging the similarity of word embeddings, it builds a candidate word set and selects the word that minimizes the predictive probability of the true class label. 

\noindent\textbf{Textbugger} \cite{Li2018TextBuggerGA} Textbugger contains both word-level and character-level perturbations by inserting, removing, swapping and substituting letters or replacing words.

\noindent\textbf{BERT-Attack} \cite{2020bertattack} BERT-Attack uses the masked language model (MLM) of BERT to replace words with other words that ﬁt the context. In addition to achieving high attack success rate, high perturbation percentage and relatively low calculation costs, BERT-Attack also ensures fluency and semanticality of adversarial samples.

\noindent\textbf{Deepwordbug} \cite{deepwordbug18} Deepwordbug designs a score system to find the rank the importance of tokens to the prediction and perturb the top $k$ important tokens by swap, substitution, deletion and insertion.

\subsection{Implementation Details}
We train 10 epochs of the models on AGNEWS and SST-2 datasets, and 20 epochs on the IMDB dataset and provide experimental results averaged on three different random seeds: 0, 1, and 2. 

For each attack, we take 1000 attack examples on the SST-2 and AGNEWS datasets, and 200 attack examples on the IMDB dataset due to the excessive number of queries. We also set a default restriction of $15\%$ on the maximum modify ratio for each attack algorithm.
Further details for implementation would be discussed in the following sections.

\subsection{Hyperparameter}
There are two main hyperparameters in our experiments: hidden dimension (hd) and $\beta$. The hidden dimension controls the dimension of the IB layer and $\beta$ controls the trade-off between better prediction performance and restriction of the information flow. We experiment with different sizes of hd ranging from 100 to 700 and different values of $\beta$ from 0.05 to 0.3. Based on model performance, we finally choose to use hd = 100 on SST-2, AGNEWS, and hd = 200 on IMDB. We used $\beta$ = 0.1 on all three datasets. All hyperparameters are chosen based on the experimental  results on the corresponding development dataset.

\begin{table*}[ht]
\small
\resizebox{\linewidth}{!}{
\begin{tabular}{l|l|c|cc|cc|cc|cc}
\hline
\hline
\multirow{2}{*}{\bf Datasets} &
  \multirow{2}{*}{\bf  Methods} &
  \multirow{2}{*}{\bf \emph{Clean\%}} &
  \multicolumn{2}{c|}{\textbf{TextFooler}} &
  \multicolumn{2}{c|}{\textbf{TextBugger}} &
  \multicolumn{2}{c|}{\textbf{BERT-Attack}} &
  \multicolumn{2}{c}{\textbf{Deepwordbug}} \\ \cline{4-11} 
                        &                  &   & \emph{Aua(Suc)\%} & \emph{\#Query} & \emph{Aua(Suc)\%} & \emph{\#Query} & \emph{Aua(Suc)\%} & \emph{\#Query}& \emph{Aua(Suc)\%} & \emph{\#Query} \\ \hline
\multirow{7}{*}{\textbf{SST-2}}   & BERT-base             & $93.2$ & $25.3(72.7)$    & $72.8$       & $35.3(61.8)$         & $43.4$       & $20.7(77.6)$        & $96.0$  & $39.2(57.6)$  & $27.5$     \\
                        & PGD         & $93.5$ & $27.9(70.2)$       & $74.6$       & $37.0(60.3) $       & $43.6$       & $22.0(76.3)$     & $96.6$  & $40.3(56.7)$ & $27.2$      \\
                        & SMART           & $\bf 94.1$ & $32.8(64.8)$        & $85.9$       & $40.9(56.1)$         & $48.2$       & $20.8(77.7) $       & $104.3$    & $45.0(51.7)$ & $29.7$   \\ 
                        & FreeLB           & $93.9$ & $29.5(68.5)$        & $73.4$       & $40.0(57.3)$         & $44.6$       & $23.7(74.7) $       & $97.0$    & $42.5(54.6)$ & $28.0$   \\ 
                        & InfoBERT         & $93.9$ & $31.5(66.2)$         & $74.1$       & $40.9(56.1) $        & $44.4$       & $25.4(72.7)$          & $99.4$ & $42.9(53.9)$ & $28.3$      \\
                        & TA-VAT         & $93.6$ & $34.6(62.6)$    & $75.4$       & $43.3(53.2) $        & $44.4$       & $26.4(71.5)$      & $99.5$    & $45.8(50.5)$ & $28.0$   \\
                        & Our Model        & $93.3$ &$37.6(59.9)$        & $104.8$       & $46.5(50.3)$          & $61.0$       & $32.9(64.9)$         & $\bf 147.0$  & $48.4(48.0)$ & $34.2$      \\  
                        & \quad+ FreeLB & $\bf 94.1$ & $\bf 40.4(56.8)$     & $\bf 106.9$       & $\bf 48.1(48.8)$        & $\bf 62.7$       & $\bf 33.3(64.5)$      & $146.8$  & $\bf 51.6(44.9)$ & $\bf 35.0$      \\ \hline 
\multirow{7}{*}{\textbf{AGNEWS}} & BERT-base             & $94.5$ & $9.1(90.4)$         & $314.1$       & $42.2(55.5)$          & $174.9$       & $13.3(86.0)$          & $414.0$    & $25.3(73.3)$ & $104.7$    \\
                        & PGD         & $94.9$ & $59.0(37.6)$     & $261.7$       & $58.8(37.8) $       & $287.3$       & $62.7(34.0)$     & $264.2$         & $65.7(30.6)$ & $254.8$ \\
                        & SMART           & $94.4$ & $54.4(42.3)$        & $155.9$       & $60.1(36.3)$         & $102.2$       & $37.8(59.9) $       & $241.0$    & $61.2(35.1)$ & $62.9$   \\ 
                        & FreeLB           & $\bf 94.7$ & $13.6(85.7)$         & $343.4$       & $47.5(49.8)$         &$175.2$       & $15.9(83.2)$      & $435.6$    & $22.3(76.4)$ & $106.3$    \\
                        & InfoBERT         & $93.6$ & $65.0(29.8)$     & $173.0$       & $68.0(26.6)$       & $106.5$       & $55.0(40.7)$       & $261.3$      & $67.0(28.0)$ & $180.3$  \\
                        & TA-VAT         & $94.5$ & $56.7(40.1)$    & $264.2$       & $56.3(40.5)$       & $290.1$       & $63.1(33.5)$      & $270.6$       & $62.3(34.2)$ & $\bf 260.0$ \\
                        & Our Model        & $94.2$ & $68.6(27.2)$       & $516.4$       & $70.8(24.9)$        & $319.9$       & $60.7(35.7)$       & $827.2$       & $70.0(26.0)$ & $130.8$ \\
                        & \quad+ FreeLB & $94.4$ & $\bf 70.8(25.0)$         & $\bf 521.3$       & $\bf 73.4(22.4)$          & $\bf 326.8$       & $\bf 64.0(32.4)$  & $\bf 851.5$   & $\bf 71.6(24.1)$ & $132.0$ \\ \hline
\multirow{7}{*}{\textbf{IMDB}}   & BERT-base             & $91.5$ & $0.8(99.1)$    & $610.8$       & $5.7(93.8)$        & $524.2$       & $0.1(99.8)$       & $570.6$  & $24.3(73.6)$ & $355.7$      \\
                        & PGD         & $92.6$ & $35.7(61.4)$       & $1911.8$       & $33.3(63.9) $    & $2261.9$       & $36.8(60.1)$       & $1549.4$    & $41.7(54.9)$ & $2082.7$    \\
                        & SMART         & $\bf 93.1$ & $48.2(48.1)$      & $2035.4$       & $53(43.2) $     & $1238.7$       & $23.2(75.2)$       & $2053.1$    & $58.8(36.8)$ & $571.8$    \\
                        & FreeLB           & $92.5$ & $38.3(58.1)$       & $1843.1$       & $49.7(45.7)$    &$1249.9$       & $26.7(70.1)$        & $2453.5$     & $57.5(37.1)$ & $550.5$   \\
                        & InfoBERT         & $91.9$ & $32.2(65.3)$        & $1112.4$       & $35.5(61.7)$        & $756.9$       & $24.5(73.5)$       & $1394.3$     & $44.7(51.8)$ & $465.0$   \\
                        & TA-VAT         & $92.5$ & $38.3(58.7)$      & $2230.1$       & $36.5(60.6) $     & $\bf 2864.5$       & $40.3(56.5)$       & $1754.4$    & $51.1(45.0)$ & $\bf 2258.17$    \\
                        & Our Model        & $91.5$ & $52.2(42.4)$       & $2077.7$        & $58.7(34.9)$      & $1366.0$       & $38.7(57.0)$        & $2859.2$     & $64.3(28.8)$ & $597.9$   \\
                        & \quad+ FreeLB & $92.4$ & $\bf 64.3(30.1)$          & $\bf 2293.8$       & $\bf 69.2(24.9)$    & $1513.4$       & $\bf 52.7(43.0)$      & $\bf 3356.0$    & $\bf 71.3(22.5)$ & $621.8$    \\ \hline
                     \hline
\end{tabular}
}
\centering
\caption{Accuracy achieved by our method and other competitive models both on clean data and under attacks. The number in bold denotes best performance on that dataset. \emph{Clean\%} denotes the prediction accuracy without attack. \emph{Aua\%} denotes accuracy under attack, \emph{Suc\%} denotes attack success rate and \emph{\#Query} denotes average query numbers. The average perturbed word percentage of all three attack methods are set to under 15\%. All the attack methods used in the experiment are from the implementation of TextAttack \cite{morris2020textattack}. All other baseline models used in this study are based on our own implementation. \label{tab:result}}
\end{table*}

\subsection{Results}

\begin{table*}[ht]
\small
\resizebox{\linewidth}{!}{
\begin{tabular}{l|l|c|cc|cc|cc|cc}
\hline
\hline
\multirow{2}{*}{\bf Datasets} &
  \multirow{2}{*}{\bf  Methods} &
  \multirow{2}{*}{\bf \emph{Clean\%}} &
  \multicolumn{2}{c|}{\textbf{TextFooler}} &
  \multicolumn{2}{c|}{\textbf{TextBugger}} &
  \multicolumn{2}{c|}{\textbf{BERT-Attack}} &
  \multicolumn{2}{c}{\textbf{Deepwordbug}} \\ \cline{4-11} 
                        &                  &   & \emph{Aua(Suc)\%} & \emph{\#Query} & \emph{Aua(Suc)\%} & \emph{\#Query} & \emph{Aua(Suc)\%} & \emph{\#Query}& \emph{Aua(Suc)\%} & \emph{\#Query} \\ \hline
\multirow{5}{*}{\textbf{SST-2}}   & BERT-base             & $93.2$ & $5.6(94.0)$       & $89.2$       & $27.9(69.8)$      & $47.5$       & $6.2(93.3) $      & $111.7$    & $27.4(70.3)$ & $29.0$    \\
                        & PGD         & $93.5$ & $6.7(92.8)$     & $92.6$       & $30.3(67.5) $     & $47.8$       & $7.5(91.9)$      & $114.3$      & $25.6(68.2)$ & $28.7$  \\
                        & SMART           & $\bf 94.1$ & $12.0(87.1)$        & $107.8$       & $33.0(64.6)$         & $53.7$       & $8.6(90.8) $       & $123.3$    & $35.1(62.4)$ & $31.2$   \\ 
                        & FreeLB           & $93.9$ & $8.1(91.4)$       & $95.4$       & $32.0(65.9)$       & $49.5$       & $9.2(90.2) $       & $118.6$     & $31.9(65.9)$ & $29.5$   \\
                        & InfoBERT         & $93.9$ & $9.5(89.8)$     & $99.3$       & $32.8 (64.8)$     & $49.6$       & $10.9(88.3)$      & $126.1$   & $32.8(64.7)$ & $29.8$     \\
                        & TA-VAT         & $93.6$ & $14.5(84.3)$       & $115.2$       & $34.9(62.3) $      & $51.5$       & $11.5(87.6)$       & $135.4$   & $35.0(62.2)$ & $29.8$     \\
                        & Our Model        & $93.3$ &$21.1(77.5)$        & $125.7$       & $39.8(57.5)$        & $68.0$       & $20.3(78.3)$       & $167.9$     & $39.1(58.1)$ & $35.9$   \\
                        & \quad+ FreeLB & $\bf 94.1$ & $\bf 23.3(75.1)$        & $\bf 129.8$       & $\bf 42.7(54.4)$         & $\bf 70.0$       & $\bf 21.3(77.3)$      & $\bf 169.0$   & $\bf 41.7(55.7)$ & $\bf 36.3$     \\ \hline
                    \hline
\end{tabular}
}
\centering
\caption{Accuracy achieved by our method and other competitors on both clean data and adversarial examples. The average perturbed word percentage of all four attack methods are not constrained. For the DeepWordBug method, the edit distance is constrained to no more than five. \label{tab:result_noconstrain}}
\end{table*}

Table \ref{tab:result} reports the detailed results of our experiment. 
As shown in Table \ref{tab:result}, our model suffers little or even no performance drop in clean accuracy on all of the three datasets. On the IMDB and AGNEWS datasets, our model achieves around the same clean accuracy as the baseline model, while on the SST-2 dataset, clean accuracy of our model in fact outperforms the baseline model by a small margin.
This implies that even with an added information bottleneck layer, the proposed method still ensures sufficient information flowing through the information bottleneck for the model to make accurate predictions.

Besides the clean performance, Table \ref{tab:result} also provides concrete experimental results on the robustness of all the models under four different types of attacks.
The robustness accuracy result shows  our model not just demonstrates significant improvement in adversarial robustness compared to BERT-base model, but is also very competitive with other defense methods under both word-level and character-level attacks. In particular, our model outperforms all baseline models by a great margin under the attack of TextFooler and TextBugger on IMDB dataset. 

Furthermore, combining our method with FreeLB (a kind of adversarial training method) can further improve models' adversarial robustness. 
We achieve the highest adversarial accuracy under all circumstances in our experiments and only suffer a small drop in clean accuracy compared to the original FreeLB method. This implies that adversarial training methods and our approach are complementary to each other.

We also evaluate the accuracy of models under adversarial attack methods whose constraint is relaxed. Specifically, the maximum modify ratio is not constrained for all attack methods, which means that relatively stronger adversarial examples can be generated.
The result in Table \ref{tab:result_noconstrain} shows that our method can still outperform all the baseline methods in this setting.

\section{Discussion}
In this section, we study how the implementation of IB layer affects the model's robustness. Furthermore, we seek to find a reasonable explanation for this effect.
\subsection{Quantitative Analysis}
\begin{figure*}[htbp]
    \centering
    \includegraphics[scale=0.17]{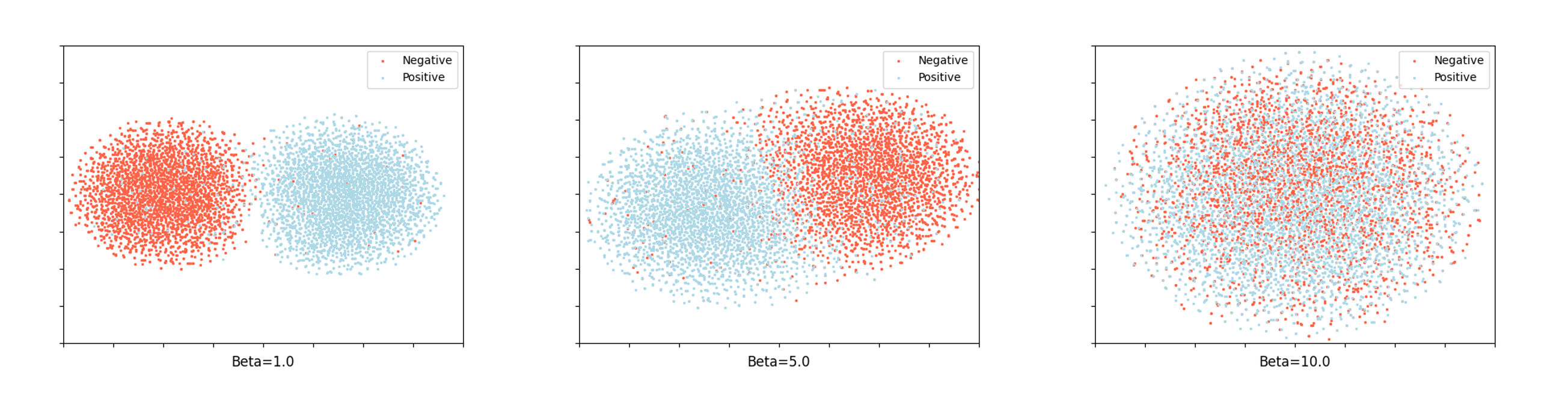}
    \caption{t-SNE visualization of our model under different $\beta$. Each marker in the figure denotes different sample. This series of figures (from left to right) shows a progression from moderate compression to too much compression. As the $\beta$ increases, the boundary between the two classes gradually disappears.}
    \label{tsne}
\end{figure*}
First, we discuss the effect of the two hyperparameters in our model: hidden dimension (hd) and $\beta$. The size of the hidden dimension controls the dimension of the information bottleneck layer, thus limiting the amount of total information that flows through the information bottleneck.
\begin{figure}[ht]
    \centering
    \includegraphics[width=7.5cm]{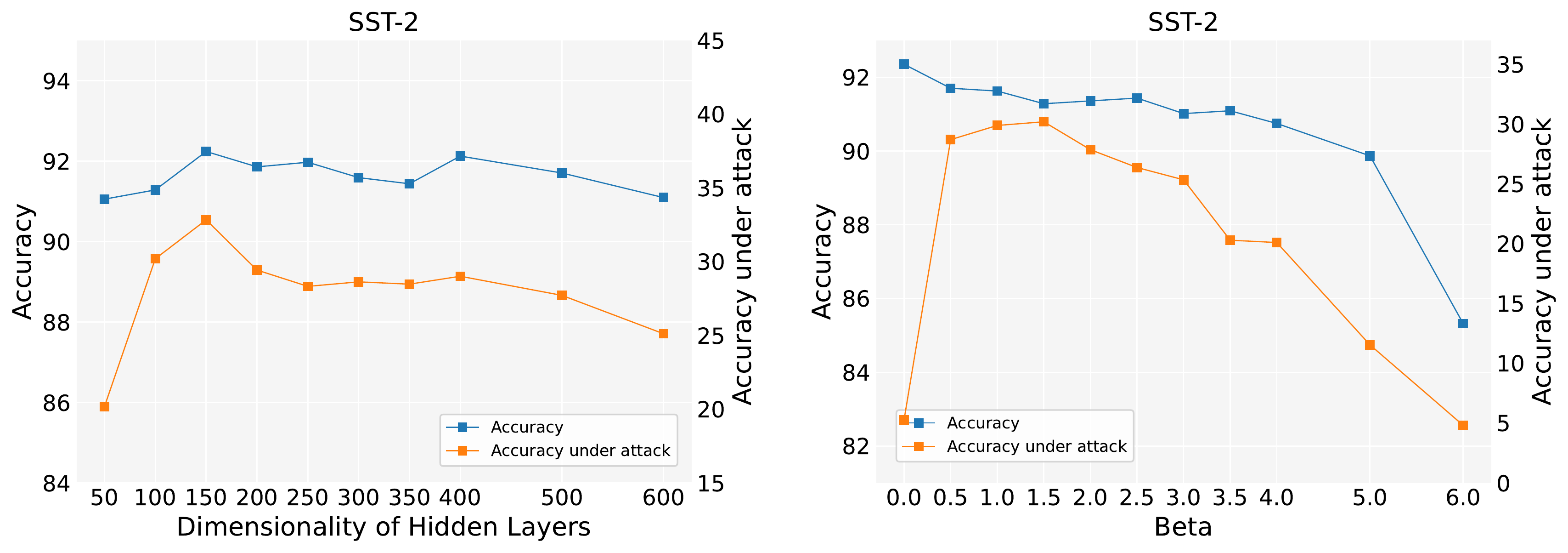}
    \caption{Performance change under different hyperparameter choices. The y-axis on the left denotes clean accuracy, while the y-axis on the right denotes accuracy under attack, using TextFooler as the attack method. The x-axis denotes different hyperparameters for our IB layer, from left to right, shows the performance of our model under different dimensionality of hidden layers and $\beta$. We fixed the hidden dimension of IB layer at 100 when choosing different values of $\beta$.}
    \label{hyper}
\end{figure}
We test the impact of 10 different hidden dimension sizes ranging from 50 to 600 on the SST-2 dataset. As shown in Figure \ref{hyper}, a small hidden dimension size helps our model achieve better performance under adversarial attacks, while only suffering little drop in clean accuracy. This indicates that choosing small hidden dimension size works best in helping the model filter out task-unrelated information. However, when the size of the hidden dimension gets too small, we observe a significant drop in accuracy under attack. 
This may indicate that when the hidden dimension size is too small, the information bottleneck layer would compress too much such that task-related information is also filtered out. 

We also test the influence of different values of $\beta$s from 0 to 6. Recall that the hyperparameter $\beta$ controls the trade off between better prediction performance and restriction of the information flow through the bottleneck. Therefore, adjusting the value of $\beta$ could also help with controlling the amount of information that flows through the information bottleneck layer. Specifically, a smaller $\beta$ will allow more information to flow from $\mathbf{X}$ to hidden representation $\mathbf{T}$, while a larger $\beta$ would ``narrow'' the bottleneck, allowing less information to flow through.
When $\beta$ is set to $0$, the information bottleneck layer is equal to a linear layer.
Results of experiments with different values of $\beta$ on the SST-2 dataset are shown in Figure \ref{hyper}. At first, as the value of $\beta$ increases, less information are able to flow through the bottleneck, forcing the information bottleneck to filter out non-robust features. This results in an enhancement of model robustness performance. However, the robust accuracy decrease as $\beta$ increase further. This might indicates that if the information bottleneck is too ``narrow'', some robust features would also be filtered out by the IB layer. 

We further visualize the influence of $\beta$  by using t-SNE. As shown in Figure \ref{tsne}, when $\beta$ is equal to $1.0$, the samples can be clearly divided into two clusters corresponding to their labels. As the value of $\beta$ increases, more information are filtered out, including those useful for the sentiment classification task. It can be observed from the figure that samples from the two categories become close to each other when the value of $\beta$ is $5.0$. Therefore, a small perturbation may cause the model to make a false prediction. The decision boundary becomes unclear when $\beta$ is set to $10.0$ and the clean accuracy of the model decreases significantly in this case.

\subsection{Interpretation}

As we have discussed in former sections, researchers such as \citeauthor{Tsi-19} and \citeauthor{advnotbugs-Ilyas} have proposed the idea that features contributing to deep learning tasks can be divided into robust features and non-robust features. 
In our assumption, the information bottleneck layer that we plug in after BERT output works to filter out non-robust features while retaining the robust features from all the information that flows out from BERT. By only preserving task-specific robust features, our model is able to attain an improvement in adversarial robustness, while minimizing the drop in clean accuracy performance at the same time.
\subsubsection{Significance Score}
In order to verify our assumption, we specifically conduct an experiment on the SST-2 dataset to see if our model is able to capture robust task-related features. In this experiment, a score $s_i$ is calculated for each word $x_i$ in the input sentence $X$ to measure the influence of $x_i$ when predicting the sentiment of the sentence. We denote the embedding of the input sentence $X$ as $E=\{e_1,...,e_n\}$, and $X_{\backslash x_i}=\{e_1,...,e_{i-1},0,e_{i+1},...,e_n\}$ denotes setting the embedding of word $x_i$ to zero.
The normalized significance score $s_i$ is defined as follows.
\begin{equation}
  s_i=\text{Rescale}(F_{Y}(X)-F_{Y}(X_{\backslash x_i}))
\end{equation}

\noindent where $F_{Y}(X)$ denotes the probability that our model gives a prediction of the label $Y$, and $\text{Rescale}(\cdot)$ is defined by dividing the significance score by the sum of the absolute values of significance scores of all words in the sentence.
Here, $s_i$ denotes the change in models' predictions before and after deleting word $x_i$. A positive value indicates that the word helps the model make correct predictions, and a negative value means that the word leads to incorrect predictions of the model.

We note that the SST-2 dataset provides sentiment labels for each word which are annotated manually. In order to identify sentiment words in sentences, we make use of these labels provided as the golden truth.
Since the words that are consistent with the sentimental tendency of the sentence are important for sentiment classification, we sum the significance scores of these words and denote this sum as $Sig\%$.

\begin{table}[t]
\small
\begin{tabular}{l|cc|cc}
\hline
\hline
\multirow{2}{*}{\bf Methods} & \multicolumn{2}{c|}{\bf  Attack} & \multicolumn{2}{c}{\bf Attack (filtered)} \\ \cline{2-5} 
                         & Sig\%       & Acc\%       & Sig\%            & Acc\%            \\ \hline
Baseline                 & $-7.9$          & $2.7 $        & $-3.6 $              & $7.3$              \\
Our model                & $4.6$           & $37.1$        & $23.7$               & $92.0$             \\ \hline
\hline
\end{tabular}
\centering
\caption{Accuracy achieved by our model and the baseline under attacks. $Sig\%$ denotes the sum of significance scores of words that are consistent with the whole sentence's sentiment tendency. $Acc\%$ denotes classification accuracy. ``Attack'' denotes the adversarial examples generated by the TextFooler algorithm, from which we choose two hundred sentences, indicated by ``Attack (filter)''. \label{tab:interpret}}
\end{table}
As shown in Table \ref{tab:interpret}, the baseline model is not able to correctly classify most sentences under textual adversarial attacks on the SST-2 dataset. Note that here $Sig\%$ is a negative value, meaning that the words that the baseline model considers ``harmful'' to making classification predictions in fact have a counter-effect on the model.
This means that the baseline model fails to correctly attention to important features in the input sentences.
By definition, $Sig\%$ increases when the models' classification accuracy improves, indicating their correlation.
Since our model achieves a higher $Sig\%$, it can be inferred that our model is better at extracting the robust features that are not likely to be perturbed under textual adversarial attacks.

\begin{figure}[h]
    \centering
    \includegraphics[width=0.45\textwidth]{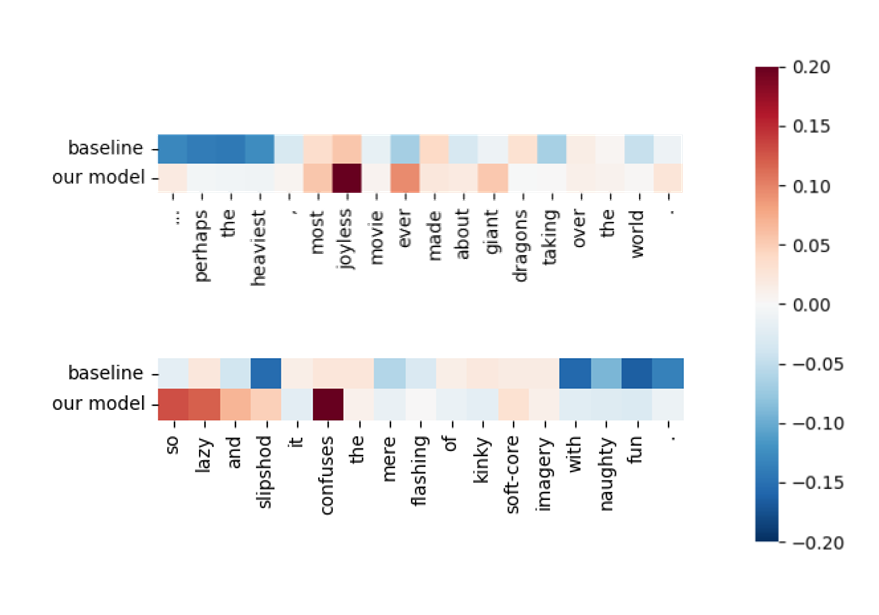}
    \caption{Illustration of each word's significance in the model's prediction process.}
    \label{fig:visual}
\end{figure}
In order to better illustrate how our model succeeds in extracting robust features, specifically for classification tasks, we visualize two examples from the SST-2 dataset in Figure \ref{fig:visual}.
In the first sentence, our model better attends to the word ``joyless'', which clearly expresses the emotional tendency in the sentence. In contrast, the attention of the baseline model is distracted by words such as ``perhaps'' and ``the'', which are almost irrelevant to the sentimental classification task. 
In the second sentence, ``slipshod'' is a sentiment word which is helpful for predicting the sentiment of the sentence. As demonstrated by the figure, our model succeeds in accurately capturing the importance of this sentiment word. The baseline model, however, failed to attend to this word and thus is unable to classify the sentence correctly.
The figure shows that our model accurately captures robust features related to important sentimental words in inputs, while the baseline model fails to do the same thing. This further validates the assumption that our model is able to extract robust task-related features for classification.

\subsubsection{Plugging Information Bottleneck into Different Layers}
\begin{figure}[h]
    \centering
    \includegraphics[width=0.4\textwidth]{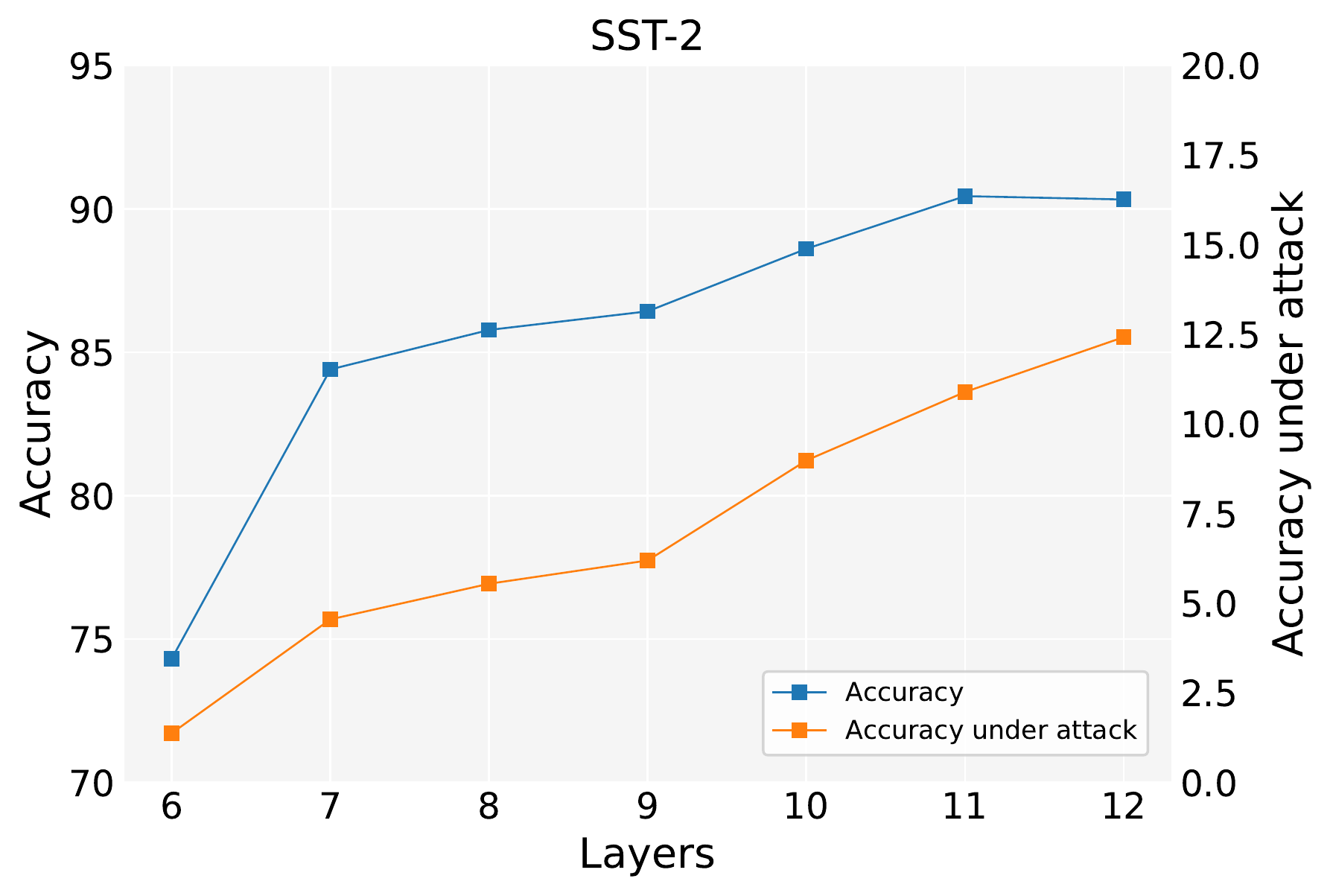}
    \caption{The performance change of our model when we add the Information Bottleneck layer after different layers of BERT model.The dimensionality of hidden layer and $\beta$ is fixed at 100 and 1.0, respectively.}
\end{figure}
Features in the deep layer model have been observed to transit from general to task-specific as the position of the layer gets higher \cite{yosinski2014transferable,howard2018universal}. More specifically, the first layers of deep neural networks may contain more general and static knowledge of the input sequences, while the higher layers contain more knowledge related to the task.
Since the information bottleneck layer is used to extract robust, task-specific features, we assume that applying it to higher layers instead of the embedding layer, which is used in InfoBERT, would be a more reasonable implementation. 
To verify this, we add the Information Bottleneck layer to the output of different layers of BERT model and calculate an additional loss(Eq.\eqref{7}) as regularization.
We train the models to minimize the total loss.
Experimental results show that as the layer we choose to apply the information bottleneck layer becomes higher, model's classification accuracy and accuracy under attack both improve, which validates our assumption.

\section{Conclusion}

We propose a novel implementation of the information bottleneck method on BERT-base models to improve its adversarial robustness. 
Our method is proven to be successful in filtering out non-robust feature and keeping task-specific robust features, thus improving the adversarial robustness of models. 
Experimental results have shown that our method outperforms four widely used defense methods across three datasets with both sentence-level and character-level attack algorithms. 
We also validate our method through a comprehensive analysis on experimental results as well as quantitative interpretation of our model's performance under adversarial attacks.

\section*{Acknowledgements}
The authors would like to thank the anonymous reviewers for their valuable comments. This work was supported by National Science Foundation of China (No. 62076068), Shanghai Municipal Science and Technology Major Project (No. 2021SHZDZX0103), and Zhangjiang Lab. Chang is supported in part by Cisco and Google. Hsieh is supported in part by NSF  IIS-2008173 and IIS-2048280.

\bibliography{anthology}

\begin{thebibliography}{67}
\expandafter\ifx\csname natexlab\endcsname\relax\def\natexlab#1{#1}\fi

\bibitem[{{Achille} and {Soatto}(2016)}]{information-dropout}
Alessandro {Achille} and Stefano {Soatto}. 2016.
\newblock \href {http://arxiv.org/abs/1611.01353} {{Information Dropout:
  Learning Optimal Representations Through Noisy Computation}}.
\newblock \emph{arXiv e-prints}, page arXiv:1611.01353.

\bibitem[{Alemi et~al.(2017)Alemi, Fischer, Dillon, and Murphy}]{DVIB017}
Alexander~A. Alemi, Ian Fischer, Joshua~V. Dillon, and Kevin Murphy. 2017.
\newblock \href {https://openreview.net/forum?id=HyxQzBceg} {Deep variational
  information bottleneck}.
\newblock In \emph{5th International Conference on Learning Representations,
  {ICLR} 2017, Toulon, France, April 24-26, 2017, Conference Track
  Proceedings}. OpenReview.net.

\bibitem[{Alzantot et~al.(2018)Alzantot, Sharma, Elgohary, Ho, Srivastava, and
  Chang}]{alzantot-etal-2018-generating}
Moustafa Alzantot, Yash Sharma, Ahmed Elgohary, Bo-Jhang Ho, Mani Srivastava,
  and Kai-Wei Chang. 2018.
\newblock \href {https://doi.org/10.18653/v1/D18-1316} {Generating natural
  language adversarial examples}.
\newblock In \emph{Proceedings of the 2018 Conference on Empirical Methods in
  Natural Language Processing}, pages 2890--2896, Brussels, Belgium.
  Association for Computational Linguistics.

\bibitem[{Chechik et~al.(2005)Chechik, Globerson, Tishby, and Weiss}]{IBG05}
Gal Chechik, Amir Globerson, Naftali Tishby, and Yair Weiss. 2005.
\newblock \href {http://jmlr.org/papers/v6/chechik05a.html} {Information
  bottleneck for gaussian variables}.
\newblock \emph{J. Mach. Learn. Res.}, 6:165--188.

\bibitem[{Chen et~al.(2020)Chen, Yu, Siqi, Tom, and Jingjing}]{iclr2020-freelb}
Zhu Chen, Gan Yu, Cheng adn~Zhe, Sun Siqi, Goldstein Tom, and Liu. Jingjing.
  2020.
\newblock {FreeLB: Enhanced Adversarial Training for Natural Language
  Understanding.}
\newblock In \emph{8th International Conference on Learning Representations,
  ICLR 2020, Addis Ababa, Ethiopia, April 26-30, 2020.}

\bibitem[{Clancey(1983)}]{c:83}
William~J. Clancey. 1983.
\newblock {Communication, Simulation, and Intelligent Agents: Implications of
  Personal Intelligent Machines for Medical Education}.
\newblock In \emph{Proceedings of the Eighth International Joint Conference on
  Artificial Intelligence {(IJCAI-83)}}, pages 556--560, Menlo Park, Calif.
  {IJCAI Organization}.

\bibitem[{Devlin et~al.(2018)Devlin, Chang, Lee, and Toutanova}]{bert}
Jacob Devlin, Ming-Wei Chang, Kenton Lee, and Kristina Toutanova. 2018.
\newblock Bert: Pre-training of deep bidirectional transformers for language
  understanding.
\newblock In \emph{Proceedings of the 2019 Conference of the North American
  Chapter of the Association for Computational Linguistics: Human Language
  Technologies.}

\bibitem[{Devlin et~al.(2019)Devlin, Chang, Lee, and
  Toutanova}]{naacl-19:Delvin}
Jacob Devlin, Ming-Wei Chang, Kenton Lee, and Kristina Toutanova. 2019.
\newblock {BERT: pre-training of deep bidirectional transformers for language
  understanding.}
\newblock In \emph{Proceedings of NAACL-HLT 2019 {(NAACL-19)}}.

\bibitem[{Ebrahimi et~al.(2018)Ebrahimi, Rao, Lowd, and
  Dou}]{ebrahimi-etal-2018-hotflip}
Javid Ebrahimi, Anyi Rao, Daniel Lowd, and Dejing Dou. 2018.
\newblock \href {https://doi.org/10.18653/v1/P18-2006} {{H}ot{F}lip: White-box
  adversarial examples for text classification}.
\newblock In \emph{Proceedings of the 56th Annual Meeting of the Association
  for Computational Linguistics (Volume 2: Short Papers)}, pages 31--36,
  Melbourne, Australia. Association for Computational Linguistics.

\bibitem[{Engelmore and Morgan(1986)}]{em:86}
Robert Engelmore and Anthony Morgan, editors. 1986.
\newblock \emph{Blackboard Systems}.
\newblock Addison-Wesley, Reading, Mass.

\bibitem[{Gao et~al.(2018{\natexlab{a}})Gao, Lanchantin, Soffa, and
  Qi}]{JiDeepWordBug18}
J.~Gao, J.~Lanchantin, M.~L. Soffa, and Y.~Qi. 2018{\natexlab{a}}.
\newblock \href {https://doi.org/10.1109/SPW.2018.00016} {Black-box generation
  of adversarial text sequences to evade deep learning classifiers}.
\newblock In \emph{2018 IEEE Security and Privacy Workshops (SPW)}, pages
  50--56.

\bibitem[{Gao et~al.(2018{\natexlab{b}})Gao, Lanchantin, Soffa, and
  Qi}]{deepwordbug18}
Ji~Gao, Jack Lanchantin, Mary~Lou Soffa, and Yanjun Qi. 2018{\natexlab{b}}.
\newblock \href {https://doi.org/10.1109/SPW.2018.00016} {Black-box generation
  of adversarial text sequences to evade deep learning classifiers}.
\newblock In \emph{2018 {IEEE} Security and Privacy Workshops, {SP} Workshops
  2018, San Francisco, CA, USA, May 24, 2018}, pages 50--56. {IEEE} Computer
  Society.

\bibitem[{Goodfellow et~al.(2014)Goodfellow, Shlens, and
  Szegedy}]{goodfellow2014}
Ian Goodfellow, Jonathon Shlens, and Christian Szegedy. 2014.
\newblock Explaining and harnessing adversarial examples.
\newblock \emph{arXiv 1412.6572}.

\bibitem[{Gulli(2004)}]{agnews}
Antonio Gulli. 2004.
\newblock Agnews.
\newblock
  \url{http://groups.di.unipi.it/˜gulli/AG_corpus_of_news_articles.html}.

\bibitem[{Han et~al.(2020)Han, Zhang, Jiang, and
  Tu}]{han-etal-2020-adversarial}
Wenjuan Han, Liwen Zhang, Yong Jiang, and Kewei Tu. 2020.
\newblock \href {https://doi.org/10.18653/v1/2020.emnlp-main.182} {Adversarial
  attack and defense of structured prediction models}.
\newblock In \emph{Proceedings of the 2020 Conference on Empirical Methods in
  Natural Language Processing (EMNLP)}, pages 2327--2338, Online. Association
  for Computational Linguistics.

\bibitem[{Hasling et~al.(1983)Hasling, Clancey, Rennels, and Test}]{hcrt:83}
Diane~Warner Hasling, William~J. Clancey, Glenn~R. Rennels, and Thomas Test.
  1983.
\newblock {Strategic Explanations in Consultation---Duplicate}.
\newblock \emph{The International Journal of Man-Machine Studies}, 20(1):3--19.

\bibitem[{Howard and Ruder(2018)}]{howard2018universal}
Jeremy Howard and Sebastian Ruder. 2018.
\newblock Universal language model fine-tuning for text classification.
\newblock In \emph{Proceedings of the 56th Annual Meeting of the Association
  for Computational Linguistics (Volume 1: Long Papers)}, pages 328--339.

\bibitem[{Huang et~al.(2019)Huang, Stanforth, Welbl, Dyer, Yogatama, Gowal,
  Dvijotham, and Kohli}]{Huang19}
Po{-}Sen Huang, Robert Stanforth, Johannes Welbl, Chris Dyer, Dani Yogatama,
  Sven Gowal, Krishnamurthy Dvijotham, and Pushmeet Kohli. 2019.
\newblock \href {https://doi.org/10.18653/v1/D19-1419} {Achieving verified
  robustness to symbol substitutions via interval bound propagation}.
\newblock In \emph{Proceedings of the 2019 Conference on Empirical Methods in
  Natural Language Processing and the 9th International Joint Conference on
  Natural Language Processing, {EMNLP-IJCNLP} 2019, Hong Kong, China, November
  3-7, 2019}, pages 4081--4091. Association for Computational Linguistics.

\bibitem[{Ilyas et~al.(2019)Ilyas, Santurkar, Tsipras, Engstrom, Tran, and
  Madry}]{advnotbugs-Ilyas}
Andrew Ilyas, Shibani Santurkar, Dimitris Tsipras, Logan Engstrom, Brandon
  Tran, and Aleksander Madry. 2019.
\newblock Adversarial examples are not bugs, they are features.
\newblock \emph{arXiv:1905.02175}.

\bibitem[{Iyyer et~al.(2018)Iyyer, Wieting, Gimpel, and
  Zettlemoyer}]{iyyer-etal-2018-adversarial}
Mohit Iyyer, John Wieting, Kevin Gimpel, and Luke Zettlemoyer. 2018.
\newblock \href {https://doi.org/10.18653/v1/N18-1170} {Adversarial example
  generation with syntactically controlled paraphrase networks}.
\newblock In \emph{Proceedings of the 2018 Conference of the North {A}merican
  Chapter of the Association for Computational Linguistics: Human Language
  Technologies, Volume 1 (Long Papers)}, pages 1875--1885, New Orleans,
  Louisiana. Association for Computational Linguistics.

\bibitem[{Jia and Liang(2017)}]{jia-liang-2017-adversarial}
Robin Jia and Percy Liang. 2017.
\newblock \href {https://doi.org/10.18653/v1/D17-1215} {Adversarial examples
  for evaluating reading comprehension systems}.
\newblock In \emph{Proceedings of the 2017 Conference on Empirical Methods in
  Natural Language Processing}, pages 2021--2031, Copenhagen, Denmark.
  Association for Computational Linguistics.

\bibitem[{Jia et~al.(2019)Jia, Raghunathan, G{\"o}ksel, and
  Liang}]{jia-etal-2019-certified}
Robin Jia, Aditi Raghunathan, Kerem G{\"o}ksel, and Percy Liang. 2019.
\newblock \href {https://doi.org/10.18653/v1/D19-1423} {Certified robustness to
  adversarial word substitutions}.
\newblock In \emph{Proceedings of the 2019 Conference on Empirical Methods in
  Natural Language Processing and the 9th International Joint Conference on
  Natural Language Processing (EMNLP-IJCNLP)}, pages 4129--4142, Hong Kong,
  China. Association for Computational Linguistics.

\bibitem[{Jiang et~al.(2020{\natexlab{a}})Jiang, He, Chen, Liu, Gao, and
  Zhao}]{acl2020-smart}
Haoming Jiang, Pengcheng He, Weizhu Chen, Xiaodong Liu, Jianfeng Gao, and Tuo
  Zhao. 2020{\natexlab{a}}.
\newblock Smart: Robust and efficient fine-tuning for pre-trained natural
  language models through principled regularized optimization.
\newblock In \emph{Proceedings of the 58th Annual Meeting of the Association
  for Computational Linguistics}, pages 2177--2190.

\bibitem[{Jiang et~al.(2020{\natexlab{b}})Jiang, He, Chen, Liu, Gao, and
  Zhao}]{jiang-etal-2020-smart}
Haoming Jiang, Pengcheng He, Weizhu Chen, Xiaodong Liu, Jianfeng Gao, and Tuo
  Zhao. 2020{\natexlab{b}}.
\newblock \href {https://doi.org/10.18653/v1/2020.acl-main.197} {{SMART}:
  Robust and efficient fine-tuning for pre-trained natural language models
  through principled regularized optimization}.
\newblock In \emph{Proceedings of the 58th Annual Meeting of the Association
  for Computational Linguistics}, pages 2177--2190, Online. Association for
  Computational Linguistics.

\bibitem[{Jin et~al.(2019)Jin, Jin, Zhou, and Szolovits}]{jin2019textfooler}
Di~Jin, Zhijing Jin, Joey~Tianyi Zhou, and Peter Szolovits. 2019.
\newblock Is bert really robust? natural language attack on text classification
  and entailment.
\newblock \emph{arXiv preprint arXiv:1907.11932}.

\bibitem[{Jin et~al.(2020)Jin, Jin, Zhou, and Szolovits}]{aaai-20:Jin}
Di~Jin, Zhijing Jin, Joey~Tianyi Zhou, and Peter Szolovits. 2020.
\newblock {Is BERT really robust? A strong baseline for natural language attack
  on text classification and entailment}.
\newblock In \emph{AAAI {(AAAI-20)}}.

\bibitem[{Kang et~al.(2018)Kang, Khot, Sabharwal, and
  Hovy}]{kang-etal-2018-adventure}
Dongyeop Kang, Tushar Khot, Ashish Sabharwal, and Eduard Hovy. 2018.
\newblock \href {https://doi.org/10.18653/v1/P18-1225} {{A}dv{E}ntu{R}e:
  Adversarial training for textual entailment with knowledge-guided examples}.
\newblock In \emph{Proceedings of the 56th Annual Meeting of the Association
  for Computational Linguistics (Volume 1: Long Papers)}, pages 2418--2428,
  Melbourne, Australia. Association for Computational Linguistics.

\bibitem[{Kurakin et~al.(2017)Kurakin, Goodfellow, and
  Bengio}]{DBLP:conf/iclr/KurakinGB17a}
Alexey Kurakin, Ian~J. Goodfellow, and Samy Bengio. 2017.
\newblock \href {https://openreview.net/forum?id=HJGU3Rodl} {Adversarial
  examples in the physical world}.
\newblock In \emph{5th International Conference on Learning Representations,
  {ICLR} 2017, Toulon, France, April 24-26, 2017, Workshop Track Proceedings}.
  OpenReview.net.

\bibitem[{Li et~al.(2018)Li, Ji, Du, Li, and Wang}]{Li2018TextBuggerGA}
Jinfeng Li, Shouling Ji, Tianyu Du, Bo~Li, and Ting Wang. 2018.
\newblock Textbugger: Generating adversarial text against real-world
  applications.
\newblock \emph{CoRR}, abs/1812.05271.

\bibitem[{Li et~al.(2019)Li, Ji, Du, Li, and Wang}]{ndss2019:li}
Jinfeng Li, Shouling Ji, Tianyu Du, Bo~Li, and Ting Wang. 2019.
\newblock {TextBugger: Generating Adversarial Text Against Real-world
  Applications.}
\newblock In \emph{NDSS 2019.}

\bibitem[{Li et~al.(2020{\natexlab{a}})Li, Ma, Guo, Xue, and
  Qiu}]{arxiv2020:li}
Linyang Li, Ruotian Ma, Qipeng Guo, Xiangyang Xue, and Xipeng Qiu.
  2020{\natexlab{a}}.
\newblock {BERT-ATTACK: Adversarial Attack Against BERT Using BERT}.
\newblock In \emph{arXiv:2004.09984 (2020)}.

\bibitem[{Li et~al.(2020{\natexlab{b}})Li, Ma, Guo, Xue, and
  Qiu}]{2020bertattack}
Linyang Li, Ruotian Ma, Qipeng Guo, Xiangyang Xue, and Xipeng Qiu.
  2020{\natexlab{b}}.
\newblock \href {https://doi.org/10.18653/v1/2020.emnlp-main.500} {Bert-attack:
  Adversarial attack against bert using bert}.
\newblock pages 6193--6202.

\bibitem[{Li and Qiu(2021)}]{tavat2021}
Linyang Li and Xipeng Qiu. 2021.
\newblock \href {https://ojs.aaai.org/index.php/AAAI/article/view/17022}
  {Token-aware virtual adversarial training in natural language understanding}.
\newblock In \emph{Thirty-Fifth {AAAI} Conference on Artificial Intelligence,
  {AAAI} 2021, Thirty-Third Conference on Innovative Applications of Artificial
  Intelligence, {IAAI} 2021, The Eleventh Symposium on Educational Advances in
  Artificial Intelligence, {EAAI} 2021, Virtual Event, February 2-9, 2021},
  pages 8410--8418. {AAAI} Press.

\bibitem[{Li and Eisner(2019)}]{2019emnlp-li}
Xiang Li and Jason Eisner. 2019.
\newblock Specializing word embeddings (for parsing) by information bottleneck
  (extended abstract).
\newblock In \emph{Proceedings of the 2019 Conference on Empirical Methods in
  Natural Language Processing and the 9th International Joint Conference on
  Natural Language Processing.}, pages 4687--4691.

\bibitem[{Maas et~al.(2011)Maas, Daly, Pham, Huang, Ng, and Potts}]{imdb}
Andrew Maas, Raymond Daly, Peter Pham, Dan Huang, Andrew Ng, and Christopher
  Potts. 2011.
\newblock Learning word vectors for sentiment analysis.
\newblock In \emph{Proceedings of the 49th Annual Meeting of the Association
  for Computational Linguistics: Human Language Technologies.}, pages 142--150.

\bibitem[{Maaten and Hinton(2008)}]{tsne}
L.J.P.V.D. Maaten and GE~Hinton. 2008.
\newblock Visualizing high-dimensional data using t-sne.
\newblock \emph{Journal of Machine Learning Research}, 9:2579--2605.

\bibitem[{Madry et~al.(2018)Madry, Makelov, Schmidt, Tsipras, and
  Vladu}]{iclr2018-aleksander}
Aleksander Madry, Aleksandar Makelov, Ludwig Schmidt, Dimitris Tsipras, and
  Adrian Vladu. 2018.
\newblock {Towards Deep Learning Models Resistant to Adversarial Attacks.}
\newblock In \emph{6th International Conference on Learning Representations,
  ICLR 2018, Vancouver, BC, Canada, April 30 - May 3, 2018, Conference Track
  Proceedings.}

\bibitem[{Miyato et~al.(2017)Miyato, Dai, and Goodfellow}]{Miyato17}
Takeru Miyato, Andrew~M. Dai, and Ian~J. Goodfellow. 2017.
\newblock \href {https://openreview.net/forum?id=r1X3g2\_xl} {Adversarial
  training methods for semi-supervised text classification}.
\newblock In \emph{5th International Conference on Learning Representations,
  {ICLR} 2017, Toulon, France, April 24-26, 2017, Conference Track
  Proceedings}. OpenReview.net.

\bibitem[{Morris et~al.(2020)Morris, Lifland, Yoo, Grigsby, Jin, and
  Qi}]{morris2020textattack}
John Morris, Eli Lifland, Jin~Yong Yoo, Jake Grigsby, Di~Jin, and Yanjun Qi.
  2020.
\newblock Textattack: A framework for adversarial attacks, data augmentation,
  and adversarial training in nlp.
\newblock In \emph{Proceedings of the 2020 Conference on Empirical Methods in
  Natural Language Processing: System Demonstrations}, pages 119--126.

\bibitem[{Nguyen et~al.(2015)Nguyen, Yosinski, and
  Clune}]{DBLP:conf/cvpr/NguyenYC15}
Anh~Mai Nguyen, Jason Yosinski, and Jeff Clune. 2015.
\newblock \href {https://doi.org/10.1109/CVPR.2015.7298640} {Deep neural
  networks are easily fooled: High confidence predictions for unrecognizable
  images}.
\newblock In \emph{{IEEE} Conference on Computer Vision and Pattern
  Recognition, {CVPR} 2015, Boston, MA, USA, June 7-12, 2015}, pages 427--436.
  {IEEE} Computer Society.

\bibitem[{Paranjape et~al.(2020)Paranjape, Joshi, Thickstun, Hajishirzi, and
  Zettlemoyer}]{paranjape-etal-2020-information}
Bhargavi Paranjape, Mandar Joshi, John Thickstun, Hannaneh Hajishirzi, and Luke
  Zettlemoyer. 2020.
\newblock \href {https://doi.org/10.18653/v1/2020.emnlp-main.153} {An
  information bottleneck approach for controlling conciseness in rationale
  extraction}.
\newblock In \emph{Proceedings of the 2020 Conference on Empirical Methods in
  Natural Language Processing (EMNLP)}, pages 1938--1952, Online. Association
  for Computational Linguistics.

\bibitem[{Ren et~al.(2019)Ren, Deng, He, and Che}]{ren-etal-2019-generating}
Shuhuai Ren, Yihe Deng, Kun He, and Wanxiang Che. 2019.
\newblock \href {https://doi.org/10.18653/v1/P19-1103} {Generating natural
  language adversarial examples through probability weighted word saliency}.
\newblock In \emph{Proceedings of the 57th Annual Meeting of the Association
  for Computational Linguistics}, pages 1085--1097, Florence, Italy.
  Association for Computational Linguistics.

\bibitem[{Sato et~al.(2018)Sato, Suzuki, Shindo, and Matsumoto}]{SatoSS018}
Motoki Sato, Jun Suzuki, Hiroyuki Shindo, and Yuji Matsumoto. 2018.
\newblock \href {https://doi.org/10.24963/ijcai.2018/601} {Interpretable
  adversarial perturbation in input embedding space for text}.
\newblock In \emph{Proceedings of the Twenty-Seventh International Joint
  Conference on Artificial Intelligence, {IJCAI} 2018, July 13-19, 2018,
  Stockholm, Sweden}, pages 4323--4330. ijcai.org.

\bibitem[{Shafahi et~al.(2019{\natexlab{a}})Shafahi, Huang, Studer, Feizi, and
  Goldstein}]{DBLP:conf/iclr/ShafahiHSFG19}
Ali Shafahi, W.~Ronny Huang, Christoph Studer, Soheil Feizi, and Tom Goldstein.
  2019{\natexlab{a}}.
\newblock \href {https://openreview.net/forum?id=r1lWUoA9FQ} {Are adversarial
  examples inevitable?}
\newblock In \emph{7th International Conference on Learning Representations,
  {ICLR} 2019, New Orleans, LA, USA, May 6-9, 2019}. OpenReview.net.

\bibitem[{Shafahi et~al.(2019{\natexlab{b}})Shafahi, Najibi, Ghiasi, Xu,
  Dickerson, Studer, Davis, Taylor, and Goldstein}]{freeat}
Ali Shafahi, Mahyar Najibi, Amin Ghiasi, Zheng Xu, John Dickerson, Christoph
  Studer, Larry Davis, Gavin Taylor, and Tom Goldstein. 2019{\natexlab{b}}.
\newblock Adversarial training for free!

\bibitem[{Socher et~al.(2013)Socher, Perelygin, Wu, Chuang, Manning, Ng, and
  Potts}]{sst-2}
Richard Socher, A.~Perelygin, J.Y. Wu, J.~Chuang, C.D. Manning, A.Y. Ng, and
  C.~Potts. 2013.
\newblock Recursive deep models for semantic compositionality over a sentiment
  treebank.
\newblock In \emph{Proceedings of the 2013 Conference on Empirical Methods in
  Natural Language Processing}, volume 1631, pages 1631--1642.

\bibitem[{Szegedy et~al.(2013)Szegedy, Zaremba, Sutskever, Bruna, Erhan,
  Goodfellow, and Fergus}]{szegedy-2013-Intriguing}
Christian Szegedy, Wojciech Zaremba, Ilya Sutskever, Joan Bruna, Dumitru Erhan,
  Ian Goodfellow, and Rob Fergus. 2013.
\newblock Intriguing properties of neural networks.
\newblock \emph{arXiv preprint arXiv:1312.6199}.

\bibitem[{Tishby et~al.(2000)Tishby, Pereira, and Bialek}]{arxiv-00:Tishby}
Naftali Tishby, Fernando~C. Pereira, and William Bialek. 2000.
\newblock {The information bottleneck method}.
\newblock \emph{arXiv:physics/0004057}.

\bibitem[{Tishby and Zaslavsky(2015)}]{ieee-15:Tishby}
Naftali Tishby and Noga Zaslavsky. 2015.
\newblock {Deep Learning and the Information Bottleneck Principle}.
\newblock \emph{arXiv:1503.02406}.

\bibitem[{Tsipras et~al.(2019)Tsipras, Santurkar, Engstrom, Turner, and
  Madry}]{Tsi-19}
Dimitris Tsipras, Shibani Santurkar, Logan Engstrom, Alexander Turner, and
  Aleksander Madry. 2019.
\newblock Robustness may be at odds with accuracy.
\newblock In \emph{International Conference on Learning Representations
  (ICLR).}

\bibitem[{Vaswani et~al.(2017)Vaswani, Shazeer, Parmar, Uszkoreit, Jones,
  Gomez, Kaiser, and Polosukhin}]{attention}
Ashish Vaswani, Noam Shazeer, Niki Parmar, Jakob Uszkoreit, Llion Jones, Aidan
  Gomez, Lukasz Kaiser, and Illia Polosukhin. 2017.
\newblock Attention is all you need.
\newblock \emph{arXiv:1706.03762}.

\bibitem[{Wang et~al.(2020)Wang, Pei, boyuan Pan, Chen, Wang, and
  Li}]{emnlp-20:Wang}
Boxin Wang, Hengzhi Pei, boyuan Pan, Qian Chen, Shuohang Wang, and Bo~Li. 2020.
\newblock {T3: Tree-autoencoder constrained adversarial text generation for
  targeted attack}.
\newblock In \emph{Proceedings of the 2020 Conference on Empirical Methods in
  Natural Language Processing}.

\bibitem[{Wang et~al.(2021{\natexlab{a}})Wang, Wang, Cheng, Gan, Jia, Li, and
  Liu}]{iclr-21:Wang}
Boxin Wang, Shuohang Wang, Yu~Cheng, Zhe Gan, Ruoxi Jia, Bo~Li, and Jingjing
  Liu. 2021{\natexlab{a}}.
\newblock {INFOBERT: Improving Robustness Of Language Models From An
  Information Theoretic Perspective}.
\newblock In \emph{ICLR {(ICLR-21)}}.

\bibitem[{Wang et~al.(2021{\natexlab{b}})Wang, Wang, Cheng, Gan, Jia, Li, and
  Liu}]{wang2021infobert}
Boxin Wang, Shuohang Wang, Yu~Cheng, Zhe Gan, Ruoxi Jia, Bo~Li, and Jingjing
  Liu. 2021{\natexlab{b}}.
\newblock Infobert: Improving robustness of language models from an information
  theoretic perspective.
\newblock In \emph{International Conference on Learning Representations}.

\bibitem[{Wang et~al.(2019)Wang, Jin, and He}]{wang2019}
Xiaosen Wang, Hao Jin, and Kun He. 2019.
\newblock \href {http://arxiv.org/abs/1909.06723} {Natural language adversarial
  attacks and defenses in word level}.
\newblock \emph{CoRR}, abs/1909.06723.

\bibitem[{Wang and Wang(2020)}]{Wang20}
Zhaoyang Wang and Hongtao Wang. 2020.
\newblock \href {https://doi.org/10.1007/978-3-030-55393-7\_28} {Defense of
  word-level adversarial attacks via random substitution encoding}.
\newblock In \emph{Knowledge Science, Engineering and Management - 13th
  International Conference, {KSEM} 2020, Hangzhou, China, August 28-30, 2020,
  Proceedings, Part {II}}, volume 12275 of \emph{Lecture Notes in Computer
  Science}, pages 312--324. Springer.

\bibitem[{West et~al.(2019)West, Holtzman, Buys, and
  Choi}]{west-etal-2019-bottlesum}
Peter West, Ari Holtzman, Jan Buys, and Yejin Choi. 2019.
\newblock \href {https://doi.org/10.18653/v1/D19-1389} {{B}ottle{S}um:
  Unsupervised and self-supervised sentence summarization using the information
  bottleneck principle}.
\newblock In \emph{Proceedings of the 2019 Conference on Empirical Methods in
  Natural Language Processing and the 9th International Joint Conference on
  Natural Language Processing (EMNLP-IJCNLP)}, pages 3752--3761, Hong Kong,
  China. Association for Computational Linguistics.

\bibitem[{Wolf et~al.(2020)Wolf, Debut, Sanh, Chaumond, Delangue, Moi, Cistac,
  Rault, Louf, Funtowicz, Davison, Shleifer, Platen, Ma, Jernite, Plu, Xu,
  Scao, Gugger, and Rush}]{transformers}
Thomas Wolf, Lysandre Debut, Victor Sanh, Julien Chaumond, Clement Delangue,
  Anthony Moi, Pierric Cistac, Tim Rault, Remi Louf, Morgan Funtowicz, Joe
  Davison, Sam Shleifer, Patrick Platen, Clara Ma, Yacine Jernite, Julien Plu,
  Canwen Xu, Teven Scao, Sylvain Gugger, and Alexander Rush. 2020.
\newblock \href {https://doi.org/10.18653/v1/2020.emnlp-demos.6} {Transformers:
  State-of-the-art natural language processing}.
\newblock In \emph{Proceedings of the 2020 EMNLP (Systems Demonstrations)},
  pages 38--45.

\bibitem[{Yosinski et~al.(2014)Yosinski, Clune, Bengio, and
  Lipson}]{yosinski2014transferable}
Jason Yosinski, Jeff Clune, Yoshua Bengio, and Hod Lipson. 2014.
\newblock How transferable are features in deep neural networks?
\newblock In \emph{NIPS}.

\bibitem[{Zang et~al.(2020)Zang, Qi, Yang, Liu, Zhang, Liu, and Sun}]{PSO20}
Yuan Zang, Fanchao Qi, Chenghao Yang, Zhiyuan Liu, Meng Zhang, Qun Liu, and
  Maosong Sun. 2020.
\newblock \href {https://doi.org/10.18653/v1/2020.acl-main.540} {Word-level
  textual adversarial attacking as combinatorial optimization}.
\newblock In \emph{Proceedings of the 58th Annual Meeting of the Association
  for Computational Linguistics, {ACL} 2020, Online, July 5-10, 2020}, pages
  6066--6080. Association for Computational Linguistics.

\bibitem[{Zhang et~al.(2019{\natexlab{a}})Zhang, Tianyuan, Lu, Zhu, and
  Dong}]{yopo}
Dinghuai Zhang, Zhang Tianyuan, Yiping Lu, Zhanxing Zhu, and Bin Dong.
  2019{\natexlab{a}}.
\newblock You only propagate once: Painless adversarial training using maximal
  principle.
\newblock In \emph{33rd Conference on Neural Information Processing Systems
  (NeurIPS 2019)}.

\bibitem[{Zhang and Yang(2018)}]{zhang2018}
Dongxu Zhang and Zhichao Yang. 2018.
\newblock \href {http://arxiv.org/abs/1804.08166} {Word embedding perturbation
  for sentence classification}.
\newblock \emph{CoRR}, abs/1804.08166.

\bibitem[{Zhang et~al.(2015)Zhang, Zhao, and LeCun}]{10.5555/2969239.2969312}
Xiang Zhang, Junbo Zhao, and Yann LeCun. 2015.
\newblock Character-level convolutional networks for text classification.
\newblock In \emph{Proceedings of the 28th International Conference on Neural
  Information Processing Systems - Volume 1}, NIPS'15, page 649–657,
  Cambridge, MA, USA. MIT Press.

\bibitem[{Zhang et~al.(2019{\natexlab{b}})Zhang, Baldridge, and
  He}]{zhang-etal-2019-paws}
Yuan Zhang, Jason Baldridge, and Luheng He. 2019{\natexlab{b}}.
\newblock \href {https://doi.org/10.18653/v1/N19-1131} {{PAWS}: Paraphrase
  adversaries from word scrambling}.
\newblock In \emph{Proceedings of the 2019 Conference of the North {A}merican
  Chapter of the Association for Computational Linguistics: Human Language
  Technologies, Volume 1 (Long and Short Papers)}, pages 1298--1308,
  Minneapolis, Minnesota. Association for Computational Linguistics.

\bibitem[{Zheng et~al.(2020)Zheng, Zeng, Zhou, Hsieh, Cheng, and
  Huang}]{acl2020-zheng}
Xiaoqing Zheng, Jiehang Zeng, Yi~Zhou, Cho-Jui Hsieh, Minhao Cheng, and
  Xuanjing Huang. 2020.
\newblock {Evaluating and Enhancing the Robustness of Neural Network-based
  Dependency Parsing Models with Adversarial Examples.}
\newblock In \emph{Proceedings of the 58th Annual Meeting of the Association
  for Computational Linguistics. {(ACL-2020)}}, page 6600–6610.

\bibitem[{Zhou et~al.(2020)Zhou, Zheng, Hsieh, Chang, and Huang}]{dne2020}
Yi~Zhou, Xiaoqing Zheng, Cho-Jui Hsieh, Kai-Wei Chang, and Xuanjing. Huang.
  2020.
\newblock {Defense against Adversarial Attacks in NLP via Dirichlet
  Neighborhood Ensemble.}
\newblock \emph{arXiv preprint arXiv:2006.11627 (2020).}

\bibitem[{Zhou et~al.(2021)Zhou, Zheng, Hsieh, Chang, and
  Huang}]{zhou-etal-2021-defense}
Yi~Zhou, Xiaoqing Zheng, Cho-Jui Hsieh, Kai-Wei Chang, and Xuanjing Huang.
  2021.
\newblock \href {https://doi.org/10.18653/v1/2021.acl-long.426} {Defense
  against synonym substitution-based adversarial attacks via {D}irichlet
  neighborhood ensemble}.
\newblock In \emph{Proceedings of the 59th Annual Meeting of the Association
  for Computational Linguistics and the 11th International Joint Conference on
  Natural Language Processing (Volume 1: Long Papers)}, pages 5482--5492,
  Online. Association for Computational Linguistics.

\end{thebibliography}
\bibliographystyle{acl_natbib}

\end{document}